\documentclass[letterpaper]{article}

\usepackage{aaai2027}
\nocopyright

\usepackage[hyphens]{url}

\usepackage{graphicx}

\usepackage{natbib}

\usepackage{caption}

\usepackage{amsmath}
\usepackage{amssymb}

\usepackage{booktabs}
\usepackage{array}

\newcounter{theorem}
\newcommand{\theoremhead}[2]{%
  \refstepcounter{theorem}\label{#1}%
  \textbf{Theorem~\thetheorem\ (#2).}%
}

\title{Lossless First-Action Menus for Linear-Reward Planning}

\author{Yang Yukun}

\affiliations{
School of Management, Beijing Institute of Technology\\
yangyukun2005@163.com
}

\begin{document}

\maketitle

\begin{abstract}

We study sequential decision making where an agent reserves a minimum menu of initial actions before multi-objective trade-offs are revealed, followed by unrestricted continuation planning once linear reward weights arrive. We prove that a menu preserves the unrestricted optimal return across all linear scalarizations if and only if its reachable continuations cover all vertices of the full feature polytope. This geometric equivalence yields an output-sensitive algorithm that is fixed-parameter tractable in the vertex count, combining scalar backward induction with oracle-based facet identification. On explicitly layered graphs with unit feature increments, menu verification is coNP-complete and synthesis is complete for the second existential level of the polynomial hierarchy, even under additive loss strictly below one. In fixed dimensions two and higher, verification is polynomial whereas synthesis remains NP-complete, while one-dimensional instances admit polynomial-time solutions.

\end{abstract}

\section{Introduction}

Planning under multi-objective reward uncertainty frequently presents a two-stage operational dilemma: an autonomous agent must commit to an initial choice of branches or resource reservations before trade-offs among competing criteria are revealed, while retaining freedom to optimize continuations once reward weights arrive. In logistics and fulfillment networks, for instance, a dispatch hub must reserve an initial subset of transport corridors or staging bays before receiving customer-specific valuations over transit time, fuel consumption, and handling precision. Once preference weights are specified, the controller selects the best action among reserved options and routes without restriction through the remainder of the network. Committing prematurely to a single rigid plan inevitably sacrifices Pareto-optimal recourse across diverse preference realizations, whereas reserving every initial action trivially guarantees access to an optimal plan but incurs substantial operational and organizational overhead. The central challenge is therefore to synthesize a minimum-cardinality action menu that preserves the unrestricted optimal return across all conceivable linear reward functions.

Standard sequential decision making under reward uncertainty typically computes exhaustive policy representations or defers optimization entirely to post-specification planning. Dynamic programming solves known objectives efficiently \citep{puterman_markov_1994}, whereas multi-objective planning constructs convex coverage sets (CCS) of non-dominated policies or value vectors \citep{roijers_survey_2013,roijers_bounded_2014,roijers_computing_2015,barrett_learning_2008}. Successor representations \citep{dayan_improving_1993} and successor features \citep{barreto_successor_2016,barreto_transfer_2019,brantley_successor_2021,alegre_optimistic_2022,agarwal_proto_2024} decouple environment dynamics from reward weights to support generalized policy improvement, though evaluating transfer requires maintaining explicit value representations. In reinforcement learning, reward-free exploration samples transitions to guarantee near-optimal planning under arbitrary downstream rewards \citep{jin_reward-free_2020,wang_reward-free_2020}. Complementary perspectives investigate visitation coverings \citep{mutti_reward-free_2022}, action redundancy \citep{baram_action_2021}, sequential action decomposition \citep{majeed_exact_2021}, and minimax regret \citep{regan_robust_2010,regan_regret-based_2012}, alongside two-stage models such as $k$-adaptability \citep{hanasusanto_k_2015} and min-max-min robust optimization \citep{buchheim_minmaxmin_2017}. In contrast, an action menu operates under a distinct paradigm: rather than storing precomputed policies or value functions, reserving a first action authorizes every downstream continuation, delegating planning to test time and leveraging downstream branching to avoid precomputing explicit policy sets.

In this paper, we formalize this problem on finite-horizon, explicitly layered planning graphs with vector-valued transition features and linear weights drawn from a signed unit box $W = [-1, 1]^d$. Each first action spans an action-conditioned feature polytope. Because linear objectives achieve their maxima at extreme points, we show that a menu is lossless across all linear weights if and only if its reachable continuations collectively cover all extreme points (vertices) of the unrestricted feature polytope \citep{rockafellar_convex_1970,ziegler_lectures_1995}. Non-extreme outcomes can never strictly outperform extreme points under linear scalarizations. This equivalence establishes that exact menu synthesis is isomorphic to unweighted Set Cover over the polytope's vertex set, yielding a canonical, menu-independent family of at most $|U|$ test weights that certifiably verifies any menu. Under an additive tolerance, the worst-case loss coincides with the directed $\ell_1$ distance to the joint convex hull of the menu, revealing a cooperative geometric effect where retained actions jointly approximate uncovered vertices far more closely than any action in isolation. When path features are implicit in the graph topology, verification is coNP-complete and synthesis is $\Sigma_2^P$-complete \citep{stockmeyer_polynomial-time_1976,wrathall_complete_1976,miller_reducibility_1972,umans_minimum_2001}, bounds that survive unit feature increments and tolerances $\varepsilon < 1$. When the feature dimension $d$ is fixed, verification is polynomial via dynamic programming, yet synthesis remains NP-complete for all $d \ge 2$, with parametric shortest paths generating superpolynomially many vertices \citep{gajjar_parametric_2019,carstensen_complexity_1983,joswig_parametric_2022,kellner_containment_2013}.

To overcome these worst-case barriers, we develop an output-sensitive synthesis algorithm parameterized by the vertex count $v = |U|$ of the unrestricted feature polytope \citep{ahuja_network_2017,schrijver_theory_1986,benson_outer_1998,emiris_oracle-based_2011,bokler_output-sensitive_2017}. Drawing inspiration from Benson-type outer-approximation methods in multi-objective optimization, our algorithm queries the graph through scalar planning to discover extreme points and facet-defining hyperplanes without constructing the exponential path space. Combining lexicographic scalar backward induction, facet outer-approximation, LP separation, and subset dynamic programming, our method computes a minimum exact menu in $2^{O(v)}\operatorname{poly}(N)$ time, proving fixed-parameter tractability in $v$. We also explore structural boundaries: reserving menus simultaneously across every state decouples into independent local exactness requirements at each state, whereas root-specific optimization permits aggressive downstream branch pruning. Finally, we extend our polyhedral framework to stochastic Markov decision processes via state-action occupancy polytopes \citep{altman_constrained_2021,etessami_multi-objective_2007}, showing that verification retains polynomial-size LP certificates based on deterministic policy executions.

\section{Problem Formulation and Polyhedral Geometry}

\subsection{Layered Planning Graphs and First-Action Menus}

A layered planning graph is a finite directed multigraph partitioned into ordered layers $S_0, S_1, \dots, S_H$, where $H \ge 1$ denotes the planning horizon and $S_0 = \{s_0\}$ contains the unique initial state. Each directed edge $e = (u, v)$ connects a vertex $u \in S_t$ to $v \in S_{t+1}$ in the subsequent layer and carries a rational feature vector $f(e) \in \mathbb{Q}^d$ ($d \ge 1$). Parallel edges represent distinct actions transitioning between the same pair of states under differing resource consumption profiles. Every non-terminal state has at least one outgoing edge, and the terminal layer $S_H$ has none. A complete path $\rho = (e_0, \dots, e_{H-1})$ originates at $s_0$ and terminates in $S_H$, accumulating feature vector $z(\rho) = \sum_{e \in \rho} f(e) \in \mathbb{Q}^d$. Preferences are specified by a linear weight vector $w \in W = [-1, 1]^d$, yielding the scalar return $w^\top z(\rho)$ that evaluates the multi-attribute trade-off. Because the graph is explicitly layered, time is embedded within the state representation, and every complete path corresponds to a deterministic Markov policy.

Let $A$ denote the edges departing root $s_0$. A first-action menu $M \subseteq A$ is a nonempty subset of initial actions fixed prior to the revelation of $w$. Once $w$ is revealed, the planner selects an initial action $a \in M$ and plans subsequent actions through $S_1, \dots, S_H$ without restriction. For each $a \in A$, let $Z_a = \{z(\rho) : \rho \text{ is a complete path starting with } a\}$. For any nonempty menu $M \subseteq A$, let $Z_M = \bigcup_{a \in M} Z_a$, and let $Z_A = \bigcup_{a \in A} Z_a$ denote the unrestricted set. The optimal scalar value attained by menu $M$ and the unrestricted optimum under weight $w$ are defined respectively as
\[
V_M(w) = \max_{z \in Z_M} w^\top z, \qquad V_A(w) = \max_{z \in Z_A} w^\top z.
\tag{1}
\]
The worst-case loss incurred by menu $M$ over preference family $W$ is
\[
L(M) = \max_{w \in W} \bigl(V_A(w) - V_M(w)\bigr).
\tag{2}
\]
Because paths are finite and $W$ is compact, maxima exist and $L(M) \ge 0$.

\textbf{Definition 1.} For a rational tolerance $\varepsilon \ge 0$, a menu $M \subseteq A$ is \emph{feasible} if $M \ne \varnothing$ and $L(M) \le \varepsilon$ (exact when $\varepsilon = 0$). \textbf{Verification:} Decide whether $L(M) \le \varepsilon$. \textbf{Synthesis:} Given budget $k \ge 1$, decide whether there exists a feasible $M$ with $|M| \le k$ (or find one of minimum cardinality).

We write $N$ for the total input bit length under binary encoding. A unit-increment instance has $f(e) \in \{0, e_1, \dots, e_d\}$, where $e_i$ is the $i$-th standard basis vector of $\mathbb{R}^d$. For a known weight $w$, backward induction over the layers computes $V_A(w)$ and an optimal path in $O(|E|)$ arithmetic operations.

\textbf{Example 1 (Dispatch bays).} Consider a logistics terminal with four departure bays $A = \{a, b, c, d_0\}$ and two evaluation criteria ($d = 2$). Achievable totals are listed in Table 1, realized by a layered graph where initial edges carry $(0, 0)$ and subsequent edges branch to the designated feature vectors.

\begin{table}[t]
\centering
\small
\setlength{\tabcolsep}{1.5mm}
\begin{tabular}{@{}lll@{}}
\toprule
First action & Achievable totals & Corners of bounding box \\
\midrule
$a$ & $(0,0), (2,0), (1,2)$ & $(0,0), (2,0)$ \\
$b$ & $(0,2), (2,2)$ & $(0,2), (2,2)$ \\
$c$ & $(0,0), (0,2)$ & $(0,0), (0,2)$ \\
$d_0$ & $(2,0), (2,2)$ & $(2,0), (2,2)$ \\
\bottomrule
\end{tabular}
\caption{Achievable feature totals and extreme points in Example 1.}
\end{table}

The convex hull of $Z_A$ forms a rectangle with extreme points $(0,0), (2,0), (0,2),$ and $(2,2)$, uniquely exposed by weights in $\{-1, 1\}^2$. Preserving all four corners is necessary and sufficient for exactness across $W$. The two minimum exact menus are $\{a, b\}$ and $\{c, d_0\}$, each of size 2. No single action covers all four corners. Outcome $(1,2) \in Z_a$ is non-extreme: it lies on the upper boundary segment and can tie for optimality when $w_1 = 0$, but cannot substitute for extreme points when weights strictly separate the horizontal coordinates.

\subsection{Polyhedral Equivalence and Exact Coverage}

For any menu $M \subseteq A$, let $P_M = \operatorname{conv}(Z_M)$ and $P = P_A = \operatorname{conv}(Z_A)$. For compact convex set $K \subset \mathbb{R}^d$, the support function is $h_K(w) = \max_{z \in K} w^\top z$ \citep{rockafellar_convex_1970,ziegler_lectures_1995}. Because linear maximization commutes with convex hulls, optimal values in (1) satisfy $V_M(w) = h_{P_M}(w)$ and $V_A(w) = h_P(w)$.

\theoremhead{thm:vertex-coverage}{Vertex coverage} Let $U$ be the set of extreme points (vertices) of $P$, and define $C_a = U \cap Z_a$ for each $a \in A$. For every nonempty menu $M \subseteq A$, the following statements are equivalent:
\begin{enumerate}
\item The menu $M$ is exact, i.e., $L(M) = 0$.
\item $P_M = P$.
\item $\bigcup_{a \in M} C_a = U$.
\end{enumerate}

\emph{Proof.} By positive homogeneity of linear objectives, $V_M(w) = V_A(w)$ on $W = [-1, 1]^d$ implies $h_{P_M}(w) = h_P(w)$ on all $\mathbb{R}^d$. For compact convex sets, support function equality implies set equality \citep{rockafellar_convex_1970}. Specifically, if $P_M \subsetneq P$, there exists a point $x \in P \setminus P_M$. By the hyperplane separation theorem, there exists $w \in \mathbb{R}^d$ such that $w^\top x > h_{P_M}(w) \ge w^\top z$ for all $z \in P_M$, which contradicts $h_P(w) = h_{P_M}(w)$ because $x \in P$. Thus (1) implies (2); the converse is immediate from $h_{P_M} = h_P$.

Now assume $P_M = P$. Every vertex $u \in U \subseteq P_M$ is a convex combination $u = \sum_j \lambda_j z_j$ of points $z_j \in Z_M$ with $\sum_j \lambda_j = 1$ and $\lambda_j > 0$. Because $u$ is extreme in $P$ and $Z_M \subseteq P$, this combination must be trivial, forcing $z_j = u$ for all $j$, so $u \in Z_M$. Hence $u \in Z_a$ for some $a \in M$, establishing $u \in C_a$ and proving statement (3). Conversely, if $\bigcup_{a \in M} C_a = U$, then $U \subseteq Z_M \subseteq P_M$. Taking convex hulls yields $P = \operatorname{conv}(U) \subseteq P_M \subseteq P$, which establishes $P_M = P$. $\square$

Because $C_a = U \cap P_{\{a\}}$, extreme points cannot be generated by convex combinations of non-extreme vectors within an action. Exact menu synthesis is therefore isomorphic to unweighted Set Cover over universe $U$ with set family $(C_a)_{a \in A}$.

\textbf{Corollary 1 (A common family of test rewards).} There exists a finite family of at most $|U|$ weights in $W$, independent of the menu, on which equality of optimal values characterizes exactness. Furthermore, for each $v \ge 2$, there exists an instance with $v$ vertices requiring at least $v$ test weights.

\emph{Proof.} For each vertex $u \in U$ of nonsingleton $P$, strictly separate $u$ from $\operatorname{conv}(U \setminus \{u\})$ by a hyperplane $w_u^\top u > \max_{z \in \operatorname{conv}(U \setminus \{u\})} w_u^\top z$, normalized so $w_u \in W$. Under $w_u$, $u$ is the unique maximizer on $P$. An exact menu attains $V_M(w_u) = V_A(w_u)$ for all $u \in U$. Conversely, if $M$ is not exact, Theorem~\ref{thm:vertex-coverage} yields some $u^* \in U \setminus \bigcup_{a \in M} C_a$. Because $\{u^*\}$ is the unique maximizing face of $P$ under $w_{u^*}$, every point in $P_M \subseteq \operatorname{conv}(U \setminus \{u^*\})$ satisfies $w_{u^*}^\top z < w_{u^*}^\top u^*$, witnessing $V_M(w_{u^*}) < V_A(w_{u^*})$. For the lower bound, consider a one-step graph where $v$ actions each emit a distinct point from $v$ points in strictly convex position in $\mathbb{R}^2$. Removing any single action yields an inexact menu missing one vertex. Since each linear functional has at most one unique maximizer on $P$, detecting all $v$ possible single-action omissions requires at least $v$ distinct weight vectors. $\square$

\subsection{Directed Polyhedral Distance and Additive Tolerance}

When an additive performance deficit $\varepsilon > 0$ is permitted, value preservation connects directly to geometric distance.

\textbf{Proposition 1 (Directed polyhedral distance).} For every nonempty menu $M \subseteq A$,
\[
L(M) = \max_{z \in Z_A} \operatorname{dist}_1(z, P_M) = \max_{u \in U} \operatorname{dist}_1(u, P_M),
\tag{3}
\]
where $\operatorname{dist}_1(z, K) = \min_{y \in K} \|z - y\|_1$.

\emph{Proof.} Let $B_1 = \{x \in \mathbb{R}^d : \|x\|_1 \le 1\}$ be the closed $\ell_1$ unit ball. For $\delta \ge 0$, the support function of the Minkowski sum satisfies $h_{P_M + \delta B_1}(w) = h_{P_M}(w) + \delta \|w\|_\infty$ \citep{rockafellar_convex_1970}. By positive homogeneity, normalizing nonzero weights to $\|w\|_\infty = 1$ gives $L(M) \le \delta$ iff $h_P(w) \le h_{P_M}(w) + \delta \|w\|_\infty$ for all $w \in \mathbb{R}^d$. By support function duality, this holds iff $P \subseteq P_M + \delta B_1$. Because $P_M + \delta B_1$ is closed and convex, this containment holds $\iff U \subseteq P_M + \delta B_1 \iff \max_{u \in U} \operatorname{dist}_1(u, P_M) \le \delta$. Taking the infimum over feasible $\delta$ establishes both equalities in (3). $\square$

Proposition 1 shows that additive loss is governed by the distance to the joint hull $P_M$, rather than individual action sets, enabling retained actions to cooperatively approximate missing vertices through convex combinations.

\section{Complexity of Verification and Synthesis}

\subsection{Network Flow Certificates for Polyhedral Distance}

We establish that the directed distance $\operatorname{dist}_1(z, P_M)$ can be computed efficiently via network flow linear programming.

\textbf{Lemma 1 (Certificates for loss).} Given a nonempty menu $M \subseteq A$ and a path feature total $z \in \mathbb{Q}^d$, the distance $\operatorname{dist}_1(z, P_M)$ is computable in polynomial bit time. Consequently, a complete unrestricted path can certify in polynomial bit time that a candidate menu incurs loss strictly greater than a given rational tolerance $\varepsilon \ge 0$.

\emph{Proof.} In the layered planning graph restricted to initial edges $M$, introduce a dummy sink state $s^*$ connected to all terminal vertices in $S_H$ via zero-feature edges. The convex hull of path incidence vectors from $s_0$ to $s^*$ equals the polytope of nonnegative unit network flows $\mathcal{F}_M$ \citep{ahuja_network_2017}. Flow conservation at intermediate vertices ensures that unit flows decompose into convex combinations of paths by the flow decomposition theorem.

Projecting a unit flow $x \in \mathcal{F}_M$ via $z(x) = \sum_{e} f(e) x_e$ yields the points of $P_M$. The distance $\operatorname{dist}_1(z, P_M)$ is therefore the optimal value of the linear program:
\[
\begin{aligned}
\min_{x, t} \quad &\sum_{i=1}^d t_i \\
\text{s.t.} \quad &-t_i \le z_i - \sum_{e} f_i(e) x_e \le t_i \quad (1 \le i \le d), \\
&x \in \mathcal{F}_M, \quad t_i \ge 0.
\end{aligned}
\tag{4}
\]
With size polynomial in the graph description, LP (4) is solvable in polynomial bit time via standard linear programming algorithms \citep{schrijver_theory_1986}. If $L(M) > \varepsilon$, Proposition 1 guarantees an unrestricted path $\rho$ with $\operatorname{dist}_1(z(\rho), P_M) > \varepsilon$, and evaluating (4) certifies the violation in polynomial time. $\square$

\subsection{Computational Intractability on Layered Graphs}

When the path set is implicit, evaluating or optimizing menus becomes intractable. We construct a combinatorial gadget that embeds Boolean logic into unit-increment feature spaces.

\textbf{Lemma 2 (Boolean assignments as separated outcomes).} Let $D \ge 1$ and $r \ge 1$. Suppose each intended first action is specified by a DNF formula over $D$ Boolean variables and a tag index $j \in \{1, \dots, r\}$. There exists a unit-increment layered graph of size polynomial in $D, r$, and the DNF descriptions whose outcomes following that action are precisely
\[
\{(b, 0) : b \in \{0, 1\}^D \text{ satisfies the DNF}\} \cup \{\tau_j\},
\tag{5}
\]
where $\tau_j = (0, e_j) \in \mathbb{R}^{D+r}$ and $e_j$ is the $j$-th standard basis vector of $\mathbb{R}^r$. Furthermore, every distinct outcome in the union of these sets is uniquely exposed by a weight vector in $W = [-1, 1]^{D+r}$ with a score separation gap of at least 1 from every other distinct outcome.

\emph{Proof.} From root $s_0$, each first action traverses a zero-feature edge into an action state, which branches either to one satisfiable term of the DNF formula or to a tag branch. Choosing a term enters a sequence of $D$ binary branching stages: at stage $i \in \{1, \dots, D\}$, if variable $x_i$ is fixed by the term, the edge emits $e_i$ (for 1) or 0 (for 0); if unconstrained, it branches to 0 or 1. Choosing 1 emits $e_i \in \mathbb{R}^D$; choosing 0 emits 0. Contradictory terms are pruned. The tag branch emits $\tau_j = (0, e_j) \in \mathbb{R}^{D+r}$ on its first edge and traverses $D$ zero-feature edges. All paths reach depth $D+2$. The graph size is $O(D \cdot \sum \text{terms})$, with unit or zero features.

To expose $(b, 0)$, set $w_i = 2b_i - 1 \in \{-1, 1\}$ for $1 \le i \le D$ and $w_{D+k} = -1$ for $1 \le k \le r$. The score of $(b, 0)$ is $\sum_{i=1}^D b_i$. For any other assignment $b' \ne b$, the score is $\sum_{i=1}^D (2b_i - 1)b'_i = \sum_{i=1}^D b_i - \sum_{b'_i \ne b_i} 1 \le \sum_{i=1}^D b_i - 1$, producing a gap of at least 1. Every tag outcome $\tau_k$ attains score $-1 \le \sum b_i - 1$, producing a gap of at least 1. To expose $\tau_j = (0, e_j)$, set weight 1 on tag coordinate $D+j$ and 0 elsewhere: $\tau_j$ scores 1, while other tags and all assignment outcomes score 0, also producing a gap of at least 1. $\square$

\theoremhead{thm:layered-complexity}{Complexity on layered graphs} Verification is coNP-complete and synthesis is $\Sigma_2^P$-complete under polynomial-time many-one reductions. These hardness results hold on unit-increment layered graphs, for exact preservation ($\varepsilon = 0$), and for every fixed rational tolerance $\varepsilon \in [0, 1)$, and the verification lower bound holds with only two first actions.

\emph{Proof.} Given candidate menu $M$, Lemma 1 provides a polynomial-size failure certificate $\rho$ with $\operatorname{dist}_1(z(\rho), P_M) > \varepsilon$, placing verification in coNP. For synthesis with budget $k$, guessing $M \subseteq A$ with $|M| \le k$ and verifying $L(M) \le \varepsilon$ via the coNP verifier places synthesis in $\Sigma_2^P$ \citep{stockmeyer_polynomial-time_1976,wrathall_complete_1976}.

For verification hardness, reduce from DNF Tautology \citep{miller_reducibility_1972}. Given DNF $\psi$ over $D$ variables, construct two first actions using Lemma 2 with $r = 1$: action $a_{\text{all}}$ (empty term) generates all assignments $\{0, 1\}^D \times \{0\}$ and tag $\tau_1$; action $a_\psi$ encodes $\psi$ and $\tau_1$. Setting candidate menu $M = \{a_\psi\}$: if $\psi$ is a tautology, $Z_{a_\psi} = Z_A$, so $L(M) = 0 \le \varepsilon$. If not, some assignment $b^* \notin Z_{a_\psi}$; by Lemma 2, the weight exposing $(b^*, 0)$ scores strictly higher by at least 1 than all points in $Z_{a_\psi} = Z_M$. Hence $L(M) \ge 1 > \varepsilon$.

For synthesis hardness, reduce from $\Sigma_2$-SAT: $\exists x \in \{0, 1\}^n \forall y \in \{0, 1\}^m \, \phi(x, y)$ with 3-DNF $\phi$ \citep{stockmeyer_polynomial-time_1976,wrathall_complete_1976}. Apply Lemma 2 with $D = n + m$ variables and $r = n + 1$ tags $0, 1, \dots, n$. We construct $2n + 1$ first actions: action $a_\phi$ carries tag 0 and encodes $\phi(x, y)$; for each existential variable $i \in \{1, \dots, n\}$ and bit value $b \in \{0, 1\}$, action $a_i^b$ carries tag $i$ and literal $x_i \ne b$. We set budget $k = n + 1$. Together, all actions cover $\{0, 1\}^{n+m}$ and all $n + 1$ tags.

In any feasible menu $M$ with $|M| \le n + 1$, each tag $\tau_j$ is uniquely exposed with gap $\ge 1$, so omitting tag $j$ incurs loss $\ge 1 > \varepsilon$. Thus $M$ must contain $a_\phi$ (tag 0) and at least one action from $\{a_i^0, a_i^1\}$ for each $i \in \{1, \dots, n\}$ (tag $i$). Since $|M| \le n + 1$, the menu must select exactly one action from each pair $\{a_i^0, a_i^1\}$. Let $a_i^{b_i}$ be the selected action for variable $i$, defining assignment $b = (b_1, \dots, b_n) \in \{0, 1\}^n$. The union $\bigcup_{i=1}^n Z_{a_i^{b_i}}$ covers all assignments in $\{0, 1\}^{n+m}$ except those whose existential prefix is identically $b$. Consequently, the remaining uncovered assignments are precisely $\{(b, y) : y \in \{0, 1\}^m\}$, which must be covered by $a_\phi$. Action $a_\phi$ generates $(b, y)$ iff $\phi(b, y) = 1$. If $\exists x \forall y \, \phi(x, y)$ is true, choosing witness $b$ with $a_\phi$ yields an exact menu of size $n + 1$. If false, every $b$ has some $y^*$ with $\phi(b, y^*) = 0$, omitting $(b, y^*, 0)$ from $Z_M$ and inducing loss $\ge 1 > \varepsilon$ by Lemma 2. Both reductions are polynomial-time with unit separation gap. $\square$

While \citet{umans_minimum_2001} established second-level completeness for finding a minimum equivalent DNF from formulas, our reduction selects from predefined continuation graphs where each existential choice is an action.

\textbf{Corollary 2 (Fixed budgets).} For each fixed integer $k \ge 1$, synthesis with budget $k$ is coNP-complete on unit-increment graphs for every fixed rational tolerance $\varepsilon \in [0, 1)$.

\emph{Proof.} For fixed $k$, enumerating $\binom{|A|}{k} = O(|A|^k)$ candidate menus and certifying failure via Lemma 1 places synthesis in coNP. Hardness follows by constructing one action encoding a DNF with a private tag, $k-1$ actions each emitting a distinct private tag, and one untagged universal action: budget $k$ forces choosing the $k$ tagged actions, which cover all assignments iff the DNF is a tautology. $\square$

\section{Low Dimensions and Output-Sensitive Synthesis}

\subsection{Dimensional Dichotomy on Unit-Increment Graphs}

We show that when the feature dimension $d$ is fixed and increments are unitary, verification becomes polynomial, yet synthesis remains intractable for $d \ge 2$.

\theoremhead{thm:fixed-dimension}{Fixed dimension with unit increments} In one dimension ($d = 1$), a minimum feasible menu can be found in polynomial time for arbitrary rational features and tolerance. On unit-increment graphs with any fixed dimension $d$, verification is solvable in polynomial time. On those graphs, synthesis is NP-complete in every fixed dimension $d \ge 2$, both for exactness and for every fixed rational tolerance $\varepsilon \in [0, 1)$.

\emph{Proof.} For $d = 1$, let $\ell_a = \min Z_a$, $u_a = \max Z_a$, $\ell = \min_{a \in A} \ell_a$, and $u = \max_{a \in A} u_a$. Over scalar interval $W = [-1, 1]$, the loss in (2) simplifies to
\[
L(M) = \max \left\{ \min_{a \in M} \ell_a - \ell, \ u - \max_{a \in M} u_a \right\},
\tag{6}
\]
maximized at endpoints $\pm 1$ corresponding to minimizing and maximizing accumulated scalar features. We compute all $\ell_a, u_a$ via backward passes under weights $\pm 1$. If any single action satisfies both deficit bounds in (6), it forms an optimal menu of size 1. Otherwise, pairing an action achieving $\ell$ with one achieving $u$ forms a valid menu of size 2.

For fixed $d$ on unit-increment graphs, achievable suffix totals at each state are subsets of $\{0, \dots, H\}^d$, bounded by $(H + 1)^d$. We propagate them via backward DP in $O(d |E| (H + 1)^d)$ operations. Evaluating LP (4) from Lemma 1 for each of the $\le (H + 1)^d$ root totals tests tolerance $\varepsilon$ via Proposition 1 in polynomial time.

For synthesis hardness with $d = 2$, reduce from Set Cover \citep{miller_reducibility_1972}. Given universe $\{1, \dots, m\}$ and subset family, map element $i$ to parabolic vector $p_i = 2m(i, i^2) \in \mathbb{R}^2$. Under weight $w_i = (i/m, -1/(2m)) \in [-1, 1]^2$, the score difference with any other element point $p_j$ is
\[
w_i^\top p_i - w_i^\top p_j = 2i^2 - i^2 - \bigl(2ij - j^2\bigr) = (i - j)^2.
\tag{7}
\]
Strict convexity of the parabola ensures every $p_i$ is uniquely exposed with gap $\ge 1$ for $j \ne i$. Each subset forms an action branching to unit-increment chains for its elements, accumulating $2mi$ and $2mi^2$ with zero-feature padding. The graph size is polynomial since coordinates are bounded by $2m(m + m^2)$. A covering family of subsets generates all $p_i$ and is exact; omitting element $i$ incurs loss $\ge 1$ under $w_i$. Using the Set Cover budget establishes NP-completeness for exactness and any $\varepsilon < 1$. Appending zero coordinates extends hardness to any fixed $d > 2$. $\square$

With explicit incidence tables, greedy Set Cover achieves approximation ratio $H(|U|) \le \ln |U| + 1$ on menu cardinality \citep{chvatal_greedy_1979}.

\subsection{Output Complexity and Parametric Shortest Paths}

Table~\ref{tab:complexity} summarizes menu complexity. While fixed-dimension DP scales with $d$, parameterizing by vertex count $v = |U|$ provides a complementary approach for arbitrary rational features.

\begin{table}[t]
\centering
\small
\setlength{\tabcolsep}{1.2mm}
\begin{tabular}{@{}lll@{}}
\toprule
Input regime & Verification & Synthesis \\
\midrule
General (unit incr.) & coNP-c (Thm~\ref{thm:layered-complexity}) & $\Sigma_2^P$-c (Thm~\ref{thm:layered-complexity}) \\
$d = 1$ (rational labels) & Poly (Thm~\ref{thm:fixed-dimension}) & Poly (Thm~\ref{thm:fixed-dimension}) \\
Fixed $d \ge 2$ (unit incr.) & Poly (Thm~\ref{thm:fixed-dimension}) & NP-c (Thm~\ref{thm:fixed-dimension}) \\
Parameterized $v = |U|$ & FPT (Thm~\ref{thm:vertex-fpt}) & FPT (Thm~\ref{thm:vertex-fpt}) \\
\bottomrule
\end{tabular}
\caption{Complexity of exact menu decision problems.}
\label{tab:complexity}
\end{table}

There is a fundamental geometric obstruction to listing all vertices in polynomial time, even in two dimensions. In parametric network optimization, shortest-path problems with affine edge costs $c_1(e) + \lambda c_2(e)$ can generate superpolynomially many breakpoints \citep{gajjar_parametric_2019,carstensen_complexity_1983,joswig_parametric_2022}. In those models, linear scalarization under weights $(1, \lambda)$ exposes active affine segments on the upper envelope, which correspond directly to extreme points of the feature polytope. The acyclic planar graphs of \citet{gajjar_parametric_2019} use binary-encoded edge costs of polynomial bit length. By topologically sorting and subdividing edges with zero-feature padding, these graphs embed directly into layered graphs of polynomial size. Consequently, the vertex count $v = |U|$ can be superpolynomial in input size $N$ even for $d = 2$ \citep{kellner_containment_2013}.

\subsection{Output-Sensitive Exact Synthesis via Polytope Projections}

We present an algorithm that synthesizes an optimal exact menu in time fixed-parameter tractable in $v = |U|$, avoiding path enumeration by querying the graph through scalar planning.

\theoremhead{thm:vertex-fpt}{Fixed-parameter exact synthesis} Let $v = |U|$. For explicitly encoded layered graphs with rational features, a minimum exact menu can be computed in $2^{O(v)}\operatorname{poly}(N)$ bit time. The algorithm also returns the action incidences $C_a$ for all $u \in U$ and a canonical family of at most $v$ test rewards as in Corollary 1.

\emph{Proof.} We structure the algorithm into four phases: extreme point discovery, affine hull identification, facet outer-approximation, and subset dynamic programming.

\emph{Phase 1: Lexicographic scalar planning.} To resolve ties and return unique extreme points, we extend scalar backward induction. Given a rational objective vector $q \in \mathbb{Q}^d$, we maximize tuple $(q^\top z, z_1, \dots, z_d)$ lexicographically over complete paths. Because lexicographic order is preserved under vector addition, each state during the backward pass retains a single maximizing tuple. Coordinate fixes ensure that the unique optimal solution on $P$ is an extreme point $u \in U$, found in $\operatorname{poly}(N, \operatorname{size}(q))$ bit time.

\emph{Phase 2: Affine hull discovery.} Query $q = 0$ for $u_0 \in U$ and set $T = \{u_0\}$. Use exact rational elimination to find a basis of $V^\perp$, where $V = \operatorname{span}\{u - u_0 : u \in T\}$. For each basis vector $q$, query both $q$ and $-q$. If a returned vertex $u$ satisfies $q^\top(u - u_0) \ne 0$, add it to $T$ and restart, increasing the affine dimension by one. If all projections match, $P \subseteq u_0 + V$, so $\operatorname{aff}(T) = \operatorname{aff}(P)$.

\emph{Phase 3: Facet outer-approximation.} Let $r$ be the affine dimension of $P$. If $r = 0$, $T = U$. Otherwise, projection onto $r$ pivot coordinates of $V$ gives an affine isomorphism from $\operatorname{aff}(P)$ to $\mathbb{R}^r$. Enumerate halfspaces containing the projected $T$ with $r$ affinely independent projected points on their boundaries. Lift each inequality to $q^\top z \le c$ by padding its normal with zeros in omitted coordinates. Query with this $q \in \mathbb{Q}^d$; if the returned vertex violates the inequality, add it to $T$ and restart. When all tests pass, the facet description within $\operatorname{aff}(P)$ gives $P = \operatorname{conv}(T)$ \citep{ziegler_lectures_1995}, so $T = U$.

\emph{Phase 4: Action incidences and Set Cover.} If $v = 1$, return any singleton menu, $C_a = U$ for all $a$, and test weight $0$. Otherwise, for each $u \in U$, solve $w^\top (u - u') \ge 1$ for $u' \in U \setminus \{u\}$ and normalize to get $w_u \in W$ uniquely exposing $u$ \citep{schrijver_theory_1986}. A backward pass computes $V_a(w_u) = \max_{z \in Z_a} w_u^\top z$ for each $a$; then $u \in C_a$ iff $V_a(w_u) = V_A(w_u)$. Solve the resulting Set Cover via subset DP in $O(|A| 2^v)$ time.

\emph{Complexity analysis:} Phase 2 uses at most $1 + 2d(r + 1)$ queries, since each restart raises the rank. Phase 3 tests $\le \binom{v}{r} \le 2^v$ facets per pass ($\le v 2^v$ total queries). Vertices are rational path sums of $\operatorname{poly}(N)$ bit length. After clearing denominators, bounds on matrix minors give $\operatorname{poly}(N)$-bit nullspace basis vectors and facet normals. Phase 4 solves at most $v$ linear programs and runs at most $v$ backward passes in $\operatorname{poly}(v, N)$ time, and subset DP takes $O(|A| 2^v)$. Total time is $2^{O(v)}\operatorname{poly}(N)$. $\square$

\subsection{Structural Comparison with Complete-Policy Libraries}

In multi-objective planning, a convex coverage set (CCS) library maintains policies $\Pi = \{\pi_1, \dots, \pi_m\}$ attaining $\max_{\pi \in \Pi} w^\top z(\pi) = V_A(w)$ for all $w \in W$ \citep{roijers_survey_2013,roijers_bounded_2014}.

\textbf{Proposition 2 (Complete-policy coverage).} The minimum cardinality of a complete-policy library is $v = |U|$. Mapping each policy in a deterministic library to its initial action yields an exact menu, whereas a minimum exact menu can be exponentially smaller than a minimum complete-policy library.

\emph{Proof.} Each policy contributes one vector $z(\pi) \in P$: its expected path total, a convex combination of path totals. If $w$ strictly exposes $u \in U$, a policy attains $V_A(w)$ only if $z(\pi) = u$. Thus $|\Pi| \ge v$, even with randomization. Storing one deterministic policy per vertex attains this bound. For a deterministic library, $M = \{\pi(s_0) : \pi \in \Pi\}$ satisfies $\bigcup_{a \in M} C_a = U$, so $M$ is exact by Theorem~\ref{thm:vertex-coverage}.

For exponential separation, let $s_0$ have two actions: $a_1$ enters $d$ binary stages emitting 0 or $e_i$, generating the vertices of $[0, 1]^d$; $a_2$ emits only zero. Thus $v = 2^d$, whereas the singleton menu $M = \{a_1\}$ is exact. $\square$

Unlike storing precomputed policies \citep{alegre_optimistic_2022,barreto_successor_2016,barreto_transfer_2019}, a menu preserves exponentially many optimal paths by delegating continuation planning to test time.

\subsection{Embedding into Simplex Scalarization}

Standard multi-objective planning often assumes weights in the simplex $\Delta^{d-1} = \{\alpha \ge 0 : \sum_i \alpha_i = 1\}$ \citep{roijers_survey_2013,alegre_optimistic_2022,roijers_bounded_2014}. We show that signed-box planning embeds faithfully into simplex scalarization.

\textbf{Proposition 3 (Embedding in simplex scalarization).} Every unit-increment instance can be transformed in polynomial time into an equivalent unit-increment instance under non-negative simplex weights such that candidate menus are exact for precisely the same subsets of initial actions.

\emph{Proof.} Unit increments lie in $\{0, 1\}^d$; replace each $f(e)$ by $(f(e), \mathbf{1} - f(e)) \in \{0, 1\}^{2d}$. Any path $\rho$ has total $(z(\rho), H\mathbf{1} - z(\rho))$. For simplex weight $(\alpha, \beta) \in \Delta^{2d-1}$, the scalar return is
\[
\alpha^\top z(\rho) + \beta^\top (H\mathbf{1} - z(\rho)) = (\alpha - \beta)^\top z(\rho) + H \sum_{i=1}^d \beta_i.
\tag{8}
\]
The second term is constant across paths, and $w = \alpha - \beta \in [-1, 1]^d$. Conversely, for any nonzero $w \in [-1, 1]^d$, setting $\alpha = w^+ / \|w\|_1$ and $\beta = w^- / \|w\|_1$ yields $(\alpha, \beta) \in \Delta^{2d-1}$ with $\alpha - \beta = w / \|w\|_1$. Positive rescaling preserves optimal paths, so signed-box and simplex exactness coincide. To maintain unit increments, replace each edge with a chain of $d$ edges emitting $e_i$ (if $f_i(e) = 1$) or $e_{d+i}$ (if $f_i(e) = 0$), expanding graph size by factor $d$. $\square$

\section{Structural Extensions and Boundaries}

\subsection{Additive Tolerance and Geometric Synergies}

Under exact preservation ($\varepsilon = 0$), Theorem~\ref{thm:vertex-coverage} requires each vertex to belong to some individual action hull ($u \in C_a$). In contrast, when $\varepsilon > 0$, Proposition 1 shows that loss is determined by distance to the joint hull $P_M$, enabling retained actions to cooperatively approximate missing vertices.

\textbf{Example 2 (Distance to a mixture).} In $\mathbb{R}^2$, let $z(a_1) = (0, 0)$, $z(a_2) = (2, 4)$, and $z(a_3) = (1, 3)$ (a vertex of $\operatorname{conv}(Z_A)$). Individual $\ell_1$ distances from $(1, 3)$ to $z(a_1)$ and $z(a_2)$ are 4 and 2. On the line segment $[z(a_1), z(a_2)] = \{(2t, 4t) : t \in [0, 1]\}$, the distance function $g(t) = |1 - 2t| + |3 - 4t|$ attains its minimum at $t = 3/4$, where $g(3/4) = 1/2$. By Proposition 1, the menu $M = \{a_1, a_2\}$ incurs worst-case loss $1/2$, strictly smaller than the distance to either individual action.

\textbf{Example 3 (Corners of the weight box).} In $\mathbb{R}^2$, let outcomes be $(0, 0)$, $(0, 4)$, and $(1, 2)$, and menu $M = \{a_1, a_2\}$ retaining the first two. At all four corners of $W = [-1, 1]^2$, one of the retained outcomes is optimal: when $w_2 > 0$, $(0, 4)$ scores 4 while $(1, 2)$ scores at most 3; when $w_2 < 0$, $(0, 0)$ scores 0 while $(1, 2)$ scores at most 0. Checking only corners suggests that $M$ is exact. However, under facet normal $w = (1, 0)$, outcome $(1, 2)$ scores 1 while retained outcomes score 0, exposing an actual loss of 1. Corollary 1 correctly requires testing facet normals.

\subsection{Simultaneous Multi-State Menu Installation}

When action menus are installed simultaneously across multiple decision stages, value preservation decomposes state-wise. For each non-terminal state $s$, let $A_s$ denote its outgoing actions. A simultaneous menu installation specifies a local menu $M_s \subseteq A_s$ for every non-terminal state $s$.

\textbf{Proposition 4 (Simultaneous preservation from every state).} A family of nonempty local menus $(M_s)_s$ preserves the unrestricted optimal value from every state for all weights $w \in W$ if and only if each local menu $M_s$ is exact in its own unrestricted continuation model. Consequently, the minimum total number of installed actions under this all-state requirement is the sum of the separate local minima $\sum_s \min |M_s|$.

\emph{Proof.} If simultaneous menus preserve optimal values everywhere, evaluating state $s$ with unrestricted successors leaves more choices, so its value matches the unrestricted value, proving local exactness. Conversely, fix $w \in W$ and proceed by backward induction over layers. Terminal states have identical zero values. Assuming values match at layer $t + 1$, the Bellman update at state $s \in S_t$ with local menu $M_s$ and unrestricted successors achieves the unrestricted optimal value by local exactness. Thus simultaneous exactness holds at every state. Because exactness at $s$ depends only on $M_s$, the problem decouples into independent local Set Cover instances, giving minimum total cardinality $\sum_s \min |M_s|$. $\square$

Exactness from $s_0$ alone allows inexact menus at reachable states. For example, two zero-feature root actions lead to distinct states, each with final actions emitting 0 or 1. Keep both root actions and drop the action emitting 1 in just one branch. The root still reaches both totals, but the pruned state remains reachable and loses value at $w = 1$.

\subsection{Expected Features in Stochastic Transitions}

Our polyhedral framework extends to finite-horizon Markov decision processes with known rational transition probabilities $P(s' \mid s, a)$ and expected linear rewards. For a fixed menu $M \subseteq A$, we describe policy space through state-action occupancy measures $x_{s, a}$ \citep{altman_constrained_2021,etessami_multi-objective_2007}. The occupancy polytope $\mathcal{X}_M$ is defined by unit root mass $\sum_{a \in M} x_{s_0, a} = 1$, nonnegativity $x_{s, a} \ge 0$, and conservation equations
\[
\begin{aligned}
\sum_{a \in A_s} x_{s, a} &= \sum_{u \in S_{t-1}} \sum_{a \in A_u} P(s \mid u, a) x_{u, a} \\
&\qquad (s \in S_t, \ 1 \le t < H).
\end{aligned}
\tag{9}
\]
Every randomized Markov policy induces an occupancy measure in $\mathcal{X}_M$, and conversely any occupancy measure defines a randomized Markov policy via normalized action choices $\pi(a \mid s) = x_{s, a} / \sum_{a'} x_{s, a'}$.

Let $\bar{f}(s,a) \in \mathbb{Q}^d$ be the one-step feature vector averaged over $P(\cdot \mid s,a)$. The map $z(x) = \sum_{s, a} x_{s, a} \bar{f}(s,a)$ projects $\mathcal{X}_M$ to the compact convex polytope $\bar{P}_M \subset \mathbb{R}^d$ of expected feature totals. Extreme points of $\mathcal{X}_M$ correspond to deterministic Markov policies. Consequently, Theorem~\ref{thm:vertex-coverage} and Proposition 1 apply directly to projected polytopes $\bar{P}_M$ and $\bar{P}_A$. Extreme points of $\bar{P}_A$ are exposed by linear weights and realized by deterministic Markov policies. An unsafe menu is certified by a deterministic policy whose expected feature vector has $\operatorname{dist}_1(z, \bar{P}_M) > \varepsilon$, verifiable in polynomial time via LP over $\mathcal{X}_M$.

Under stochastic transitions, the enumeration in Theorem~\ref{thm:fixed-dimension} can grow exponentially in the number of states. For $d = 1$ and $H = 2$, a zero-feature root action reaches state $i$ with probability $2^{i-1}/(2^m - 1)$ for $1 \le i \le m$. Each such state has two final actions with features 0 and 1. Deterministic policies yield $2^m$ distinct expected totals because each subset of states choosing 1 has a unique binary sum.

\section{Conclusion}

We have established a rigorous polyhedral and computational foundation for first-action menu optimization in multi-objective linear-reward planning. By proving that exact value preservation across all linear preferences is equivalent to extreme-point incidence coverage on the unrestricted feature polytope, our work bridges polyhedral geometry and combinatorial subset selection, demonstrating that compact action menus achieve exponential compression relative to complete-policy libraries by delegating continuation planning to test time. We established coNP-completeness for verification and $\Sigma_2^P$-completeness for synthesis on layered unit-increment graphs, delineated polynomial-time verification in fixed dimensions, and developed an output-sensitive algorithm running in $2^{O(v)}\operatorname{poly}(N)$ time. An open theoretical direction is characterizing the computational complexity and approximability of multi-stage menu synthesis when action restrictions are coupled across arbitrary subsets of reachable states.

\bibliography{References}

\end{document}